\documentclass[runningheads]{llncs}
\usepackage{graphicx}
\usepackage{comment}
\usepackage{amsmath,amssymb} % define this before the line numbering.
\usepackage{color}
\usepackage{float}
\usepackage{cite}
\usepackage[colorlinks, citecolor=green]{hyperref}
\usepackage{tikz}
\usepackage{epsfig}
\usepackage{graphicx}
\usepackage{booktabs}
\usepackage{enumitem}
\usepackage{subfigure}
\usepackage{multirow}
\usepackage{tikz}
\usepackage{comment}
\usepackage{amsmath,amssymb} % define this before the line numbering.
\usepackage{color}

\makeatletter
\def\thanks#1{\protected@xdef\@thanks{\@thanks
        \protect\footnotetext{#1}}}
\makeatother
\begin{document}

\pagestyle{headings}
\mainmatter
\def\ECCVSubNumber{2755}  % Insert your submission number here

\title{Face Super-Resolution Guided by \\ 3D Facial Priors} % Replace with your title

% INITIAL SUBMISSION 
%\begin{comment}
% \titlerunning{ECCV-20 submission ID \ECCVSubNumber} 
% \authorrunning{ECCV-20 submission ID \ECCVSubNumber} 
% \author{Anonymous ECCV submission}
% \institute{Paper ID \ECCVSubNumber}
%\end{comment}
%******************

% % CAMERA READY SUBMISSION
%\begin{comment}
\titlerunning{Face Super-Resolution Guided by 3D Facial Priors}
% If the paper title is too long for the running head, you can set
% an abbreviated paper title here
% \footnote[*]{*indicates corresponding author}
% Xiaobin Hu, Wenqi Ren, John LaMaster, Xiaochun Cao, Xiaoming Li, Zechao Li, bjoern menze, Wei Liu
\author{Xiaobin Hu\inst{1,2} \and
Wenqi Ren\inst{2\ast \thanks{$\ast$ indicates the corresponding author.}  } \and 
John LaMaster\inst{1} \and
Xiaochun Cao\inst{2,6} \and
Xiaoming Li\inst{3}  \and
Zechao Li\inst{4 } \and
Bjoern Menze\inst{1\dagger \thanks{$\dagger$ these authors contributed equally to this work.} } \and
Wei Liu\inst{5\dagger} 
}

% \footnote{123}
%\thanks{$\ast$ indicates corresponding author }
%\footnote
\authorrunning{Hu X.B., Ren W.Q., et al.}
%\institute[short]{\inst{1} First \and \inst{2} Second}
% \institute[short]{\inst{1} First \samelineand \inst{2} Second \samelineand \inst{3} Third}
%\institute{\inst{1} Informatics, Technische Universit\"at M\"unchen, Germany \hspace{0.1cm} \inst{2} SKLOIS, IIE, CAS \\ \inst{3} Harbin Institute of Technology  \hspace{0.1cm} \inst{4}  NJUST \hspace{0.1cm} \inst{5} Tencent AI Lab \\ \inst{6} Peng Cheng Laboratory, Cyberspace Security Research Center, China}

\institute{Informatics, Technische Universit\"at M\"unchen, Germany \hspace{0.1cm} \and SKLOIS, IIE, CAS \hspace{0.1cm} \inst{3} Harbin Institute of Technology  \hspace{0.1cm} \inst{4}  NJUST \hspace{0.1cm} \inst{5} Tencent AI Lab \\ \inst{6} Peng Cheng Laboratory, Cyberspace Security Research Center, China}

%\footnote{\dagger}
%\hspace{1cm}
%\qquad
%\end{comment}
%******************
\maketitle

\begin{abstract}
State-of-the-art face super-resolution methods employ deep convolutional neural networks to learn a mapping between low- and high-resolution facial patterns by exploring local appearance knowledge. However, most of these methods do not well exploit facial structures and identity information, and struggle to deal with facial images that exhibit large pose variations. In this paper, we propose a novel face super-resolution method that explicitly incorporates 3D facial priors which grasp the sharp facial structures. Our work is the first to explore 3D morphable knowledge based on the fusion of parametric descriptions of face attributes (e.g., identity, facial expression, texture, illumination, and face pose). Furthermore, the priors can easily be incorporated into any network and are extremely efficient in improving the performance and accelerating the convergence speed. Firstly, a 3D face rendering branch is set up to obtain 3D priors of salient facial structures and identity knowledge. Secondly, the Spatial Attention Module is used to better exploit this hierarchical information (i.e., intensity similarity, 3D facial structure, and identity content) for the super-resolution problem. Extensive experiments demonstrate that the proposed 3D priors achieve superior face super-resolution results over the state-of-the-arts.
% \dots
\end{abstract}

\section{Introduction}
Face images provide crucial clues for human observation as well as computer analysis \cite{computer_analysis2,Li_new}. However, the performance of most face image tasks, such as face recognition and facial emotion detection \cite{ed1,ed2}, degrades dramatically when the resolution of a facial image is relatively low. Consequently, face super-resolution, also known as face hallucination, was coined to restore a high-resolution face image from its low-resolution counterpart. 

Although a great influx of deep learning methods \cite{deep9,deep10,prior_yu,deep11,deep6,deep12,deep13,deep16,pami_aided,real_application2,Liu_wei} have been successfully applied in face Super-Resolution (SR) problems, super-resolving arbitrary facial images, especially at high magnification factors, is still an open and challenging problem due to the ill-posed nature of the SR problem and the difficulty in learning and integrating strong priors into a face hallucination model. 
Some researches \cite{prior_1,prior_2,Ren_ICCV,sr_identity,comparsion_1, comparsion_2} on exploiting the face priors to assist neural networks in capturing more facial details have been proposed. A face hallucination model incorporating identity priors was presented in \cite{prior_1}. However, the identity prior was extracted only from the multi-scale up-sampling results in the training procedure and therefore cannot provide extra priors to guide the network.
Yu et al.\cite{prior_2} employed facial component heatmaps to encourage the upsampling stream to generate super-resolved faces with higher-quality details, especially for large pose variations. 
Kim et al.\cite{comparsion_1} proposed a face alignment network (FAN) for landmark heatmap extraction to boost the performance of face SR.
Chen et al.\cite{comparsion_2} utilized the heatmaps and parsing maps for face SR problems. Although these 2D priors provide global component regions, these methods cannot learn the 3D reconstruction of detailed edges, illumination, and expression priors. In addition, all of these aforementioned face SR approaches ignore facial structure and identity recovery.

\begin{figure}[t]\scriptsize
	\begin{center}
		\tabcolsep 1pt
		\begin{tabular}{@{}cccccc@{}}
			\includegraphics[width=0.15\textwidth]{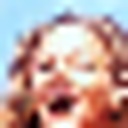}  & 
			\includegraphics[width=0.15\textwidth]{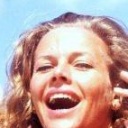}  & 
			\includegraphics[width=0.15\textwidth]{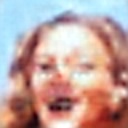}  & 
			\includegraphics[width=0.15\textwidth]{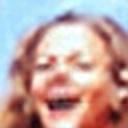}  & 
			
            \includegraphics[width=0.15\textwidth]{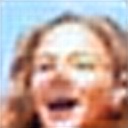} & 
            
			\includegraphics[width=0.15\textwidth]{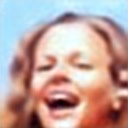}  \\
			
			(a) LR & (b) HR & (c) SRCNN & (d) SRCNN+3D & (e) VDSR & (f) VDSR+3D \vspace{1pt}\\
			PSNR/SSIM & - & 19.18/0.5553 & 21.10/0.6100 & 19.74/0.5772 & 22.44/0.6797 \vspace{1pt}\\
			\includegraphics[width=0.15\textwidth]{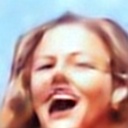}  & 
			\includegraphics[width=0.15\textwidth]{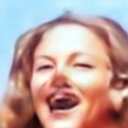} & 
			\includegraphics[width=0.15\textwidth]{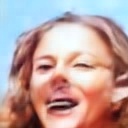}&
			\includegraphics[width=0.15\textwidth]{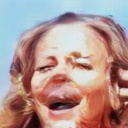}  & 
			\includegraphics[width=0.15\textwidth]{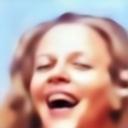} & 
			\includegraphics[width=0.15\textwidth]{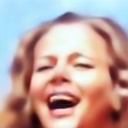} \vspace{1pt} \\
			(g) RCAN & (h) RDN & (i) Wavelet & (j) PSR-FAN & (k) FSR-Net & (l) Ours
			\\
			19.68/0.6350 & 19.81/0.6470 & 19.28/0.6232 & 19.62/0.6123& 22.17/0.6804 & \textbf{22.83/0.7172} \vspace{1pt}\\
		\end{tabular}
	\end{center}
	\vspace{-6mm}
\caption{Visual comparison with state-of-the-art face hallucination methods ($\times$8). (a) 16$\times$16 LR input. (b) 128$\times$128 HR ground-truth. (c) Super-Resolution Convolutional Neural Network (SRCNN) \cite{network6}. (d) SRCNN incorporating our 3D facial priors.  (e) Very Deep Super-Resolution Network (VDSR) \cite{network15}. (f) VDSR incorporating our 3D facial priors. (g) Very Deep Residual Channel Attention Network (RCAN) \cite{network_zhang}. (h) Residual Dense Network (RDN) \cite{network_rdn}. (i) Wavelet-based CNN for Multi-scale Face Super-Resolution (Wavelet-SRNet) \cite{network_wave}. (j) Progressive Face Super-Resolution using the facial landmark (PSR-FAN) \cite{comparsion_1}.  (k) End-to-End Learning Face Super-Resolution with Facial Priors (FSRNet) \cite{comparsion_2}. (l) Our proposed method by embedding the 3D facial priors into the Spatial Attention Module (SAM3D).}
%\label{fig:fig1}
	\vspace{-4mm}
	\label{fig:fig1}
\end{figure}

%new add
In contrast to the aforementioned approaches, we propose a novel face super resolution method by exploiting 3D facial priors to grasp sharp face structures and identity knowledge.
%which blind by 2D priors. 
Firstly, a deep 3D face reconstruction branch is set up to explicitly obtain 3D face render priors which facilitate the face super-resolution branch. 
Specifically, the 3D facial priors 
%are generated by the ResNet-50 network \cite{resnet} and 3D Morphable Model (3DMM). It 
contain rich hierarchical features, such as low-level (e.g., sharp edge and illumination) and perception level (e.g., identity) information.
Then, a spatial attention module is employed to adaptively integrate the 3D facial prior into the network, in which we employ a spatial feature transform (SFT) \cite{sft} to generate affine transformation parameters for spatial feature modulation. Afterwards, it encourages the network to learn the spatial inter-dependencies of features between 3D facial priors and input images after adding the attention module into the network. As shown in Figure \ref{fig:fig1}, by embedding the 3D rendered face priors, our algorithm generates clearer and sharper facial structures without any ghosting artifacts compared with other 2D prior-based methods.

The main contributions of this paper are:
%\vspace{-2mm}
\begin{itemize}%[leftmargin=*]
    \item[$\bullet$] A novel face SR model is proposed by explicitly exploiting facial structure in the form of facial prior estimation. The estimated 3D facial prior provides not only spatial information of facial components but also their 3D visibility information, which is ignored by the pixel-level content and 2D priors (e.g., landmark heatmaps and parsing maps).
    \item[$\bullet$] To well adapt to the 3D reconstruction of low-resolution face images, we present a new skin-aware loss function projecting the constructed 3D coefficients onto the rendered images. In addition, we use a feature fusion-based network to better extract and integrate the face rendered priors by employing a spatial attention module.
    \item[$\bullet$] Our proposed 3D facial prior has a high flexibility because its modular structure allows for easy plug-in of any SR methods (e.g., SRCNN and VDSR). We qualitatively and quantitatively evaluate the proposed algorithm on multi-scale face super-resolution, especially at very low input resolutions. The proposed network achieves better SR criteria and superior visual quality compared to state-of-the-art face SR methods.
\end{itemize}

%\vspace{-2mm}
\section{Related Work}
Face hallucination relates closely to the natural image super-resolution problem. In this section, we discuss recent research on super-resolution and face hallucination to illustrate the necessary context for our work.

%\vspace{-2mm}
{\flushleft \textbf{Super-Resolution Neural Networks.}} Recently, neural networks have demonstrated a remarkable capability to improve SR results. Since the pioneering network \cite{network6} demonstrates the effectiveness of CNN to learn the mapping between LR and HR pairs, a lot of CNN architectures have been proposed for SR \cite{network7,network26,network18,network10,network16,network28}. Most of the existing high-performance SR networks have residual blocks \cite{network15} to go deeper in the network architecture, and achieve better performance. EDSR \cite{network20} improved the performance by removing unnecessary batch normalization layers in residual blocks. A residual dense network (RDN) \cite{network_rdn} was proposed to exploit the hierarchical features from all the convolutional layers. Zhang et al.\cite{network_zhang} proposed the very deep residual channel attention networks (RCAN) to discard abundant low-frequency information which hinders the representational ability of CNNs. Wang et al.\cite{sft} used a spatial feature transform layer to introduce the semantic prior as an additional input of the SR network. Huang et al.\cite{network_wave} presented a wavelet-based CNN approach that can ultra-resolve a very low-resolution face image in a unified framework. Lian et al.\cite{real_application} proposed a Feature-Guided Super-Resolution Generative Adversarial Network (FG-SRGAN) for unpaired image super-resolution. However, these networks require a lot of time to train the massive parameters to obtain good results. In our work, we largely decrease the training parameters, but still achieve superior performance in the SR criteria (SSIM and PSNR) and visible quality. 

%\vspace{-2mm}
{\flushleft \textbf{Facial Prior Knowledge.}} Exploiting facial priors in face hallucination, such as spatial configuration of facial components \cite{deblure_semantic}, is the key factor that differentiates it from generic super-resolution tasks. There are some face SR methods that use facial prior knowledge to super-resolve LR faces. Wang and Tang \cite{prior_wang} learned subspaces from LR and HR face images, and then reconstructed an HR output from the PCA coefficients of the LR input. Liu et al.\cite{prior_liu} set up a Markov Random Field (MRF) to reduce ghosting artifacts because of the misalignments in LR images. However, these methods are prone to generating severe artifacts, especially with large pose variations and misalignments in LR images. 
Yu and Porikli \cite{prior_yu} interweaved multiple spatial transformer networks \cite{prior_jader} with the deconvolutional layers to handle unaligned LR faces. Dahl et al.\cite{deep13} leveraged the framework of PixelCNN \cite{prior_35} to super-resolve very low-resolution faces. Zhu et al.\cite{deep6} presented a cascade bi-network, dubbed CBN, to localize LR facial components first and then upsample the facial components; however, CBN may produce ghosting faces when localization errors occur. Recently, Yu et al.\cite{prior_2} used a multi-task convolutional neural network (CNN) to incorporate structural information of faces. Grm et al.\cite{prior_1} built a face recognition model that acts as identity priors for the super-resolution network during training. Yu et al.\cite{comparsion_2} constructed an end-to-end SR network to incorporate the facial landmark heatmaps and parsing maps. Kim et al.\cite{comparsion_1} proposed a compressed version of the face alignment network (FAN) to obtain landmark heatmaps for the SR network in a progressive method. 
However, existing face SR algorithms only employ 2D priors without considering high-dimensional information (3D). In this paper, we exploit the 3D face reconstruction branch to extract the 3D facial structure, detailed edges, illumination, and identity priors to guide face image super-resolution.
%Furthermore, we recover these priors in an explicit way.

%\vspace{-2mm}
{\flushleft \textbf{3D Face Reconstruction.}} The 3D shapes of facial images can be restored from unconstrained 2D images by the 3D face reconstruction. In this paper, we employ the 3D Morphable Model (3DMM) \cite{3dm_2,3dm_6,3dm_3} based on the fusion of parametric descriptions of face attributes (e.g., gender, identity, and distinctiveness) to reconstruct the 3D facial priors. The 3D reconstructed face will inherit the facial features and present the clear and sharp facial components.
{\flushleft{Closest to ours is the work of Ren et al.\cite{Ren_ICCV} which utilizes the 3D priors in the task of face video deblurring. Our method differs in several important ways. First, instead of simple priors concatenation, we employ the Spatial Feature Transform Block to incorporate the 3D priors in the intermediate layer by adaptively adjusting the modulation parameter pair. Specifically, the outputs of the SFT layer are adaptively controlled by the modulation parameter pair by applying an affine transformation spatially to each intermediate feature map. Second, the attention mechanism is embeded into the network as a guide to bias the allocation of most informative components and the interdependency between the 3D priors and input.}}
%\vspace{-2mm}
\section{The Proposed Method}
\begin{figure}[t]
\centering
\includegraphics[width=0.9\textwidth]{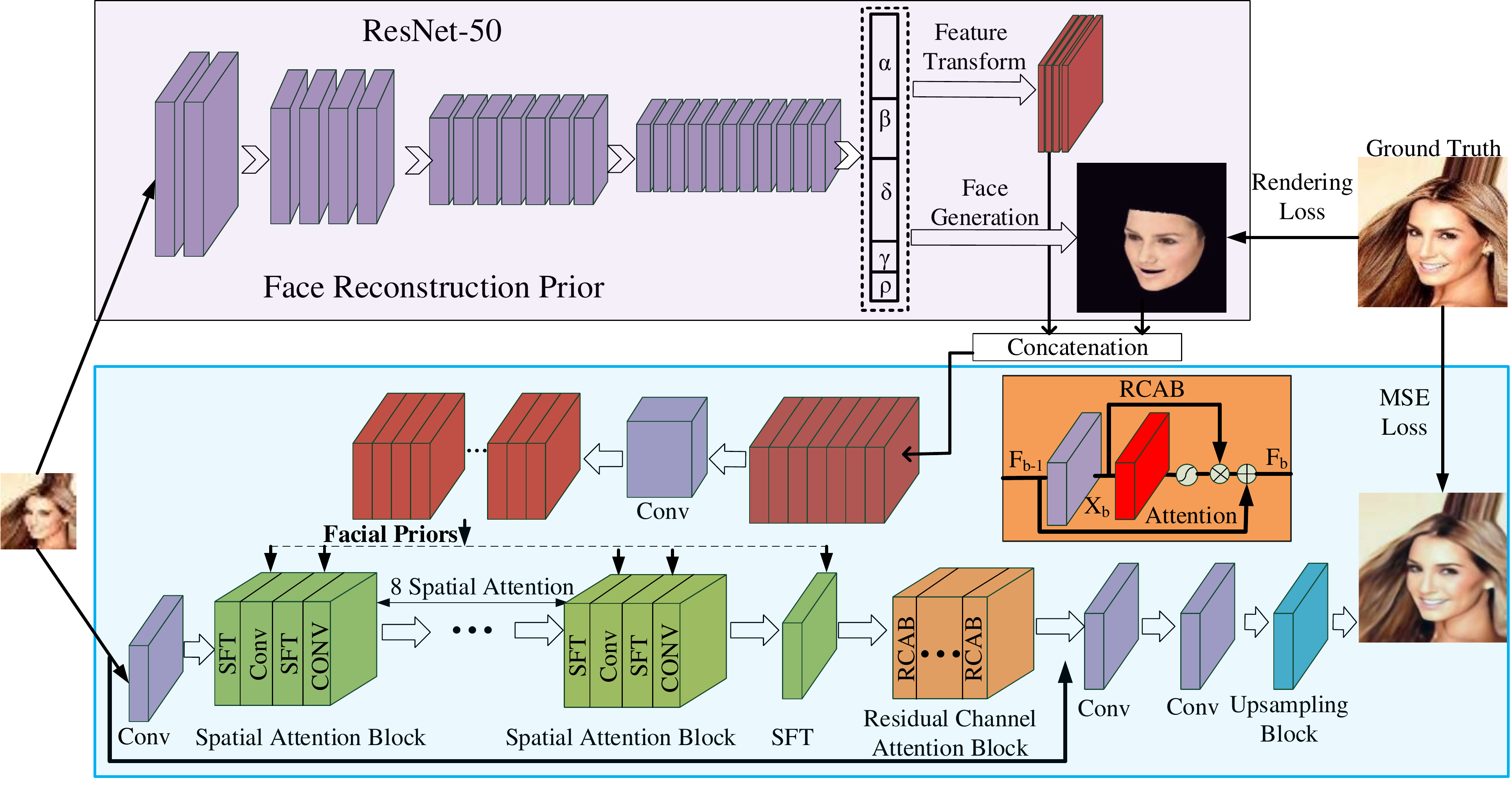}
\vspace{-4mm}
\caption{The proposed face super-resolution architecture. Our model consists of two branches: the top block is a ResNet-50 Network to extract the 3D facial coefficients and restore a sharp face rendered structure. The bottom block is dedicated to face super-resolution guided by the facial coefficients and rendered sharp face structures which are concatenated by the Spatial Feature Transform (SFT) layer.}
\label{fig:fig2}
\vspace{-4mm}
\end{figure}

%\vspace{-2mm}
The proposed face super-resolution framework presented in Figure \ref{fig:fig2} consists of two branches: the 3D rendering network to extract the facial prior and the spatial attention module aiming to exploit the prior for the face super-resolution problem.
Given a low-resolution face image, we first use the 3D rendering branch to extract the 3D face coefficients. Then a high-resolution rendered image is generated using the 3D coefficients and regarded as the high-resolution facial prior which facilitates the face super-resolving process in the spatial attention module.

\begin{figure}[t]\scriptsize
\centering
\includegraphics[width=1\textwidth]{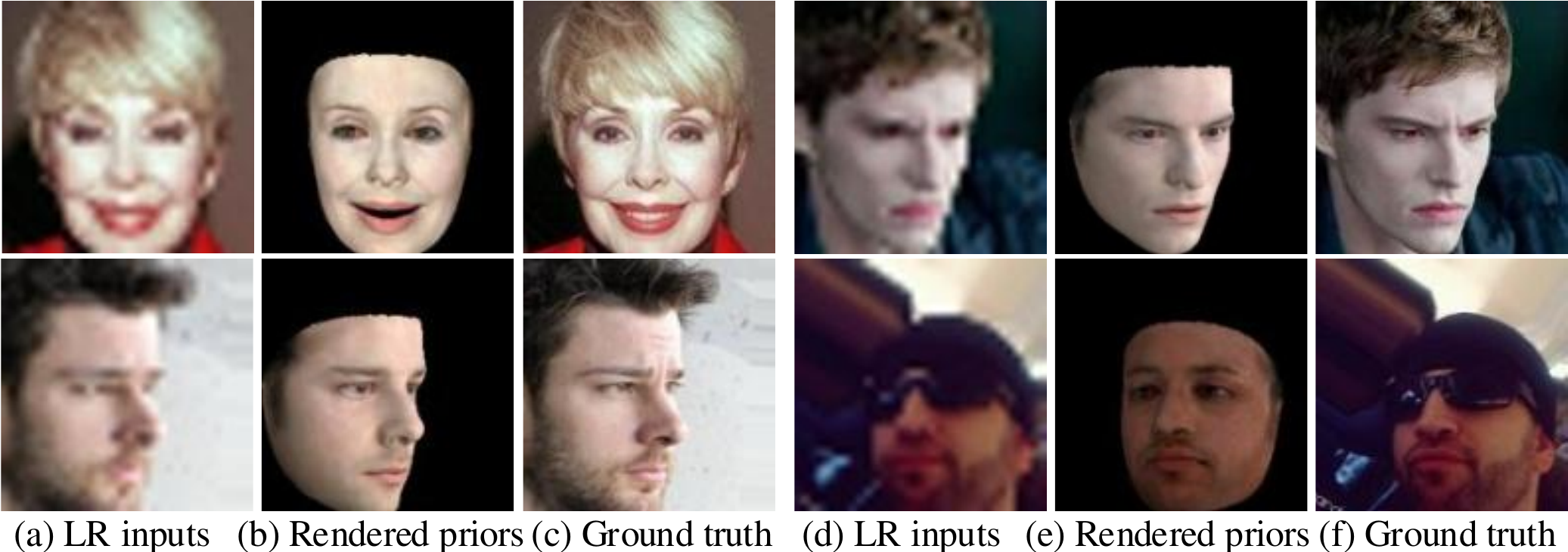}
\vspace{-6mm}
\caption{The rendered priors from our method. (a) and (d) low-resolution inputs. (b) and (e) our rendered face structures. (c) and (f) ground-truths. As shown, the reconstructed facial structures provide clear spatial locations and sharp visualization of facial components even for large pose variations (e.g., left and right facial pose positions) and partial occlusions.}\label{fig:fig3}
\vspace{-6mm}
\end{figure}

%\vspace{-2mm}
\subsection{Motivations and Advantages of 3D Facial Priors}
Existing face SR algorithms only employ 2D priors without considering high dimensional information (3D). The 3D morphable facial priors are the main novelty of this work and are completely different from recently related 2D prior works (\textit{e.g.,} the parsing maps and facial landmark heatmaps by FSRNet \cite{comparsion_2} and the landmark heatmap extraction by FAN \cite{comparsion_1}). The 3D coefficients contain abundant hierarchical knowledge, such as identity, facial expression, texture, illumination, and face pose. Furthermore, in contrast with the 2D landmark-based priors whose attentions only lie at the distinct points of facial landmarks that may lead to the facial distortions and artifacts, our 3D priors are explicit and visible, and can generate the realistic and robust HR results, greatly reducing artifacts even for large pose variations and partial occlusions.

Given low-resolution face images, the generated 3D rendered reconstructions are shown in Figure \ref{fig:fig3}. The rendered face predictions contain the clear spatial knowledge and sharp visual quality of facial components which are close to the ground-truth, even in images containing large pose variations as shown in the second row of Figure \ref{fig:fig3}.
%The 3D priors grasp very well the pose variations and skin color, and further embed pose variations into the super-resolution networks which improve the accuracy and stability in face images with large pose variations. 
Therefore, we concatenate the reconstructed face image as an additional feature in the super-resolution network. The face expression, identity, texture, the element-concatenation of illumination, and face pose are transformed into four feature maps and fed into the spatial feature transform block of the super-resolution network. 

For real-world applications of the 3D face morphable model, there are typical problems to overcome, including large pose variations and partial occlusions.
%WQ: don't forget to show it.
As shown in the supplementary material, the morphable model can generate realistic reconstructions of large pose variations, which contain faithful visual quality of facial components. The 3D model is also robust and accurately restores the rendered faces partially occluded by glasses, hair, etc. In comparison with other SR algorithms which are blind to unknown degradation types, our 3D model can robustly generate the 3D morphable priors to guide the SR branch to grasp the clear spatial knowledge and facial components, even for complicated real-world applications. Furthermore, our 3D priors can be plugged into any network and largely improve the performance of existing SR networks (\textit{e.g.,} SRCNN and VDSR demonstrated in Section 5). 
%textcolor{red}{Section \ref{}}).

%\vspace{-2mm}
\subsection{Formulation of 3D Facial Priors}
It is still a challenge for state-of-the-art edge prediction methods to acquire very sharp facial structures from low-resolution images. Therefore, a 3DMM-based model is proposed to localize the precise facial structure by generating the 3D facial images which are constructed by the 3D coefficient vector. In addition, there exist large face pose variations, such as in-plane and out-of-plane rotations. A large amount of data is needed to learn the representative features varying with the facial poses. 
To address this problem, an inspiration came from the idea that the 3DMM coefficients can analytically model the pose variations with a simple mathematical derivation \cite{3dm_3,3dm_6} and do not require a large training set. As such, we utilize a face rendering network based on ResNet-50 to regress a face coefficient vector. The output of the ResNet-50 is the representative feature vector of $\boldsymbol{x}=(\boldsymbol{\alpha},\boldsymbol{\beta},\boldsymbol{\delta},\boldsymbol{\gamma},\boldsymbol{\rho})\in\mathbb{R}^{239}$, where $\ \boldsymbol{\alpha}\in\mathbb{R}^{80},\boldsymbol{\beta}\in\mathbb{R}^{64},\boldsymbol{\delta}\in\mathbb{R}^{80},\boldsymbol{\gamma}\in\mathbb{R}^{9}, \ $and$ \ \boldsymbol{\rho}\in\mathbb{R}^{6}$ represent the identity, facial expression, texture, illumination, and face pose \cite{3dm_6}, respectively.

According to the Morphable model \cite{3dm_2}, we transform the face coefficients to a 3D shape \textbf{S} and texture \textbf{T} of the face image as
\begin{equation}
\textbf{S}=\textbf{S}(\boldsymbol{\alpha},\boldsymbol{\beta})= \overline{\textbf{S}}+\textbf{B}_{id}\boldsymbol{\alpha}+\textbf{B}_{exp}\boldsymbol{\beta},
\end{equation}
and
\begin{equation}
\textbf{T}=\textbf{T}(\boldsymbol{\delta})= \overline{\textbf{T}}+\textbf{B}_{t}\boldsymbol{\delta},
\end{equation}
where $\overline{\textbf{S}}$ and $\overline{\textbf{T}}$ are the average values of face shape and texture, respectively. $\textbf{B}_{t}$, $\textbf{B}_{id}$, and $\textbf{B}_{exp}$ denote the base vectors of texture, identity, and expression calculated by the PCA method. We set up the illumination model by assuming a Lambertian surface for faces, and estimate the scene illumination with Spherical Harmonics (SH) \cite{illumination1} to derive the illumination coefficient $\boldsymbol{\gamma}\in\mathbb{R}^{9}$. The 3D face pose $\boldsymbol{\rho} \in\mathbb{R}^{6}$ is represented by rotation $ \textbf{R}\in \rm{SO} (3)  $ and translation $ \textbf{t} \in \mathbb{R}^{3}$.

To stabilize the rendered faces, a modified $L_2$ loss function for the 3D face reconstruction is presented based on a paired training set
\begin{equation}
\ell _{r}=\frac{1}{L}\sum_{j=1}^{L}\frac{\sum_{i\in{M}}{A}^i\left \| I^{i}_{j}-R^{i}_{j}(B(\boldsymbol{x})) \right\|_{2}}{\sum_{i\in{M}}{A}^i},
\end{equation}
where $j$ is the paired image index, $L$ is the total number of training pairs, $i$ and $M$ denote the pixel index and face region, respectively, $I$ represents the sharp image, and $A$ is a skin color based attention mask obtained by training a Bayes classifier with Gaussian Mixture Models \cite{3dm_6}.
In addition, $x$ represents the LR (input) images, $B(x)$ denotes the regressed coefficients obtained by the ResNet-50 with input $x$ as input, and finally $R$ denotes the image rendered with the 3D coefficients $B(x)$.  Rendering is the process to project the constructed 3D face onto the 2D image plane with the regressed pose and illumination. We use a ResNet-50 network to regress these coefficients by modifying the last fully-connected layer to 239 neurons ( the same number of the coefficient parameters).

% For more details regarding the 3D face model and the rendering process, we refer the reader to \cite{3dm_6}.

%\vspace{-2mm}
{\flushleft \textbf{Coefficient Feature Transformation.}}
%
%After obtaining the 3D coefficients and the rendered face, 
Our 3D face priors consist of two parts: one directly from the rendered face region (\textit{i.e.,} the RGB input), and the other from the feature transformation of the coefficient parameters. The coefficient parameters  $\boldsymbol{\alpha}, \boldsymbol{\beta}, \boldsymbol{\delta}, \boldsymbol{\gamma}, \boldsymbol{\rho}$ represent the identity, facial expression, texture, illumination, and face pose priors, respectively. The coefficient feature transformation procedure is described as follows: firstly, the coefficients of identity, expression, texture, and the element-concatenation of illumination and face pose ($\boldsymbol{\gamma} +\boldsymbol{\rho}$) are reshaped to four matrices by setting extra elements to zeros. Afterwards, these four matrices are expanded to the same size as the LR images (16$\times$16 or 32$\times$32) by zero-padding, and then scaled to the interval [0,1]. Finally, the coefficient features are concatenated with the priors of the rendered face images.

%\vspace{-2mm}
\subsection{Spatial Attention Module}
%
% \textcolor{red}{To exploit the 3D face rendered priors, we propose a Spatial Attention Module (SAM) to grasp the precise locations of face components and the facial identity. The proposed SAM consists of three parts: a spatial feature transform block, an residual channel attention block, and an upscale block.}
To exploit the 3D face rendered priors, we propose a Spatial Attention Module (SAM) to grasp the precise locations of face components and the facial identity. The proposed SAM consists of three parts: a spatial feature transform block, a residual channel attention block, and an upscale block.
%
%To explore the interdependence and correlation of priors and input images between channels, the attention block is added into the spatial attention module. , 

\begin{figure}[t]
\centering
\includegraphics[width=0.7\textwidth]{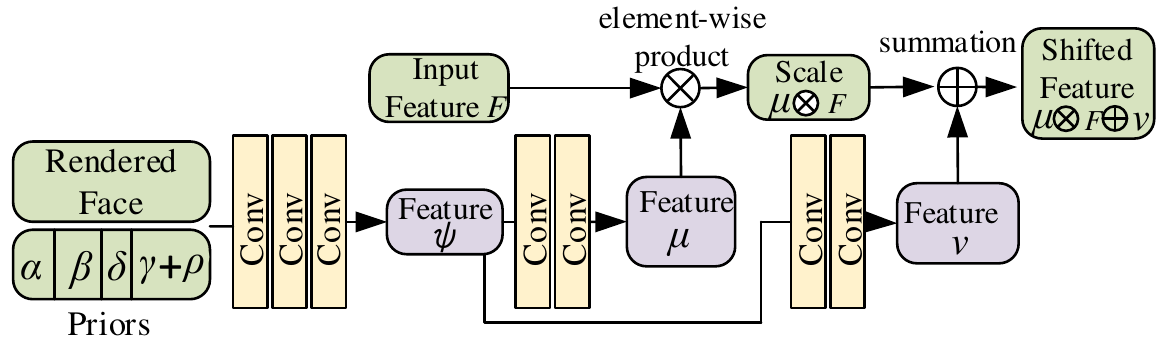}
\vspace{-4mm}
\caption{The structure of the SFT layer. The rendered faces and feature vectors are regarded as the guidance for face super-resolution.}
\label{fig:fig4}
\vspace{-2mm}
\end{figure}

%\vspace{-2mm}
{\flushleft \textbf{Spatial Feature Transform Block.}}
 The 3D face priors (rendered faces and coefficient features) are imported into the spatial attention transform block \cite{sft} after a convolutional layer. The structure of the spatial feature transform layer is shown in Figure \ref{fig:fig4}. The SFT layer learns a mapping function $\Theta$ that provides a modulation parameter pair $(\mu,\nu)$ according to the priors $\psi$, such as segmentation probability. Here, the 3D face priors are taken as the input. The outputs of the SFT layer are adaptively controlled by the modulation parameter pair by applying an affine transformation spatially to each intermediate feature map. Specifically, the intermediate transformation parameters $(\mu,\nu)$  are derived from the priors $\psi$ by the mapping function:
\begin{equation}
(\mu,\nu)=\Theta(\psi),
\end{equation}
The intermediate feature maps are modified by scaling and shifting feature maps according to the transformation parameters:

\begin{equation}
\boldsymbol{SFT}(\boldsymbol{F}|\boldsymbol{\mu},\boldsymbol{\nu})=\boldsymbol{\mu}\otimes\boldsymbol{F}+\boldsymbol{\nu},
\end{equation}
where $\boldsymbol{F}$ denotes the feature maps, and $\otimes$ indicates element-wise multiplication. At this step, the SFT layer implements the spatial-wise transformation.

{\flushleft \textbf{Residual Channel Attention Block.}}
An attention mechanism can be viewed as a guide to bias the allocation of available processing resources towards the most informative components of the input \cite{attention_12}. Consequently, the channel mechanism is presented to explore the most informative components and the inter-dependency between the channels. Inspired by the residual channel network \cite{network_zhang}, the attention mechanism is composed of a series of residual channel attention blocks (RCAB) shown in Figure \ref{fig:fig2}. For the $b$-th block, the output $\boldsymbol{F_b}$ of RCAB is obtained by:

\begin{equation}
\boldsymbol{F_b}=\boldsymbol{F_{b-1}}+C_b(\boldsymbol{X_{b}})\cdot\boldsymbol{X_{b}},
\end{equation}
where $C_b$ denotes the channel attention function. $\boldsymbol{F_{b-1}}$ is the block's input, and $\boldsymbol{X_{b}}$ is calculated by two stacked convolutional layers. The upscale block is progressive deconvolutional layers (also known as transposed convolution).
%
% \textcolor{red}{There is no introduction of the upscale block. Even its simple, you need to explain how to upsample the features and how to generate the final output.}

% Inspired by the integration of channel attention and residual blocks, we ensemble a series of residual channel attention blocks. 
% For the $b$-th block, the output $F_b$ of RCAB is obtained by:
% \begin{equation}
% \boldsymbol{F_b}=\boldsymbol{F_{b-1}}+C_b(\boldsymbol{X_{b}})\cdot\boldsymbol{X_{b}},
% \end{equation}
% where $C_b$ denotes the channel attention function. $\boldsymbol{F_{b-1}}$ is the block's input, and $\boldsymbol{X_{b}}$ is calculated by two stacked convolutional layers.
%\vspace{-2mm}
\section{Experimental Results}
%\vspace{-2mm}
%
To evaluate the performances of the proposed face super-resolution network, we qualitatively and quantitatively compare our algorithm against nine start-of-the-art super-resolution and face hallucination methods including: the Very Deep Super Resolution Network (VDSR) \cite{network15}, the Very Deep Residual Channel Attention Network (RCAN) \cite{network_zhang}, the Residual Dense Network (RDN) \cite{network_rdn}, the Super-Resolution Convolutional Neural Network (SRCNN) \cite{network6}, the Transformative Discriminative Autoencoder (TDAE) \cite{prior_yu}, the Wavelet-based CNN for Multi-scale Face Super Resolution (Wavelet-SRNet) \cite{network_wave}, the deep end-to-end trainable face SR network (FSRNet) \cite{comparsion_2}, face SR generative adversarial network (FSRGAN) \cite{comparsion_2} incorporating the 2D facial landmark heatmaps and parsing maps, and the progressive face Super Resolution network via  face alignment network (PSR-FAN) \cite{comparsion_1} using 2D landmark heatmap priors. We use the open-source implementations from the authors and train all the networks on the same dataset for a fair comparison. For simplicity, we refer to the proposed network as Spatial Attention Module guided by 3D priors, or SAM3D. In addition, to demonstrate the plug-in characteristic of the proposed 3D facial priors, we propose two models of SRCNN+3D and VDSR+3D by embedding the 3D facial prior as an extra input channel to the basic backbone of SRCNN \cite{network6} and VDSR \cite{network15}.
%. Both of them are the basic backbone of SRCNN \cite{network6} and VDSR \cite{network15} embedded with the 3D facial prior as extra RGB channel information.
%
%which is the basic VDSR model  and the other is our SR network incorporating facial priors by the Spatial Attention Module (SAM). 
The implementation code will be made available to the public. More analyses and results can be found in the supplementary material.  

\begin{figure}[t]
\centering
\includegraphics[width=1\textwidth]{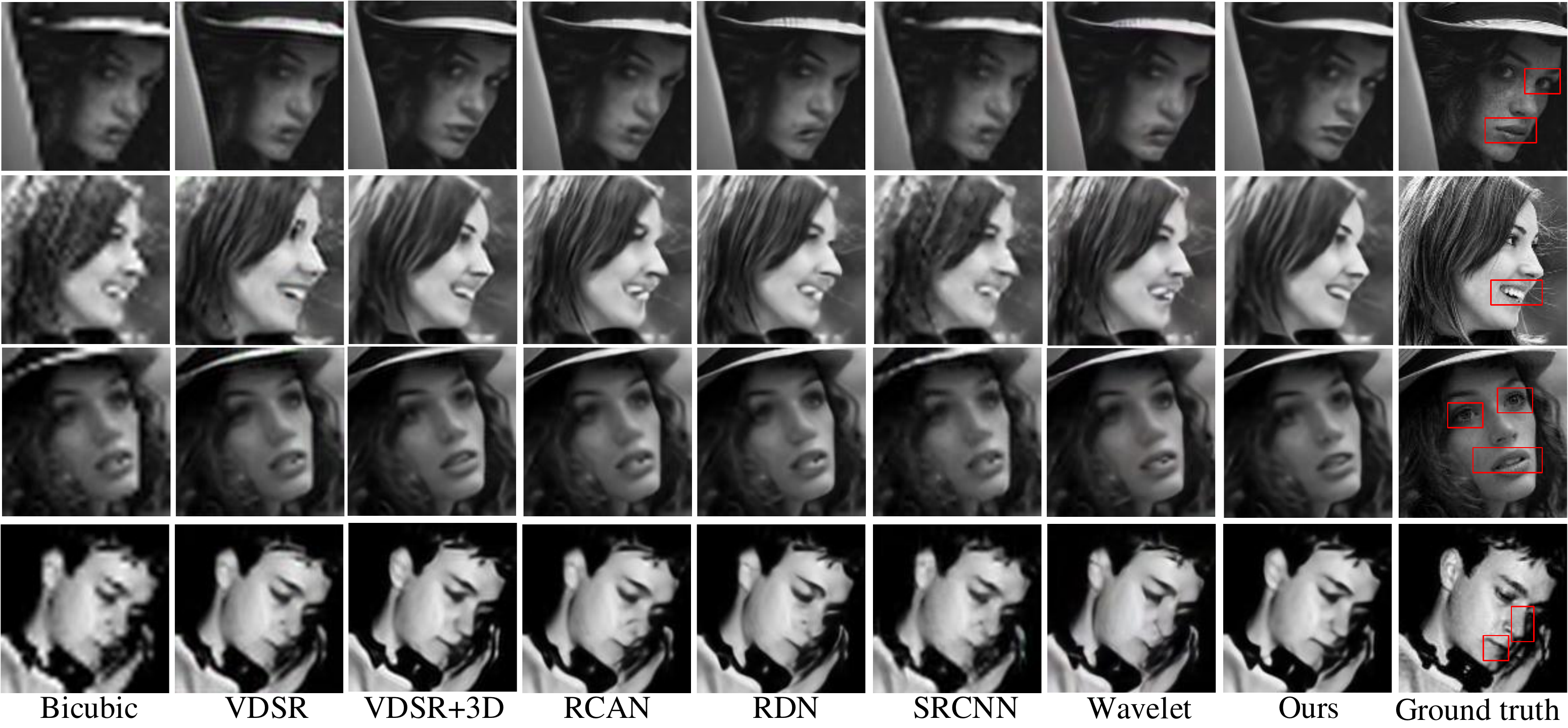}
\vspace{-6mm}
\caption{Comparison of state-of-the-art methods: magnification factors $\times$4 and the input resolution 32$\times$32. Our algorithm is able to exploit the regularity present in face regions rather than other methods. Best viewed by zooming in on the screen.}\label{fig:fig5}
%\space{-4mm}
\end{figure}
\begin{figure}[!ht]
\centering
\includegraphics[width=0.85\textwidth]{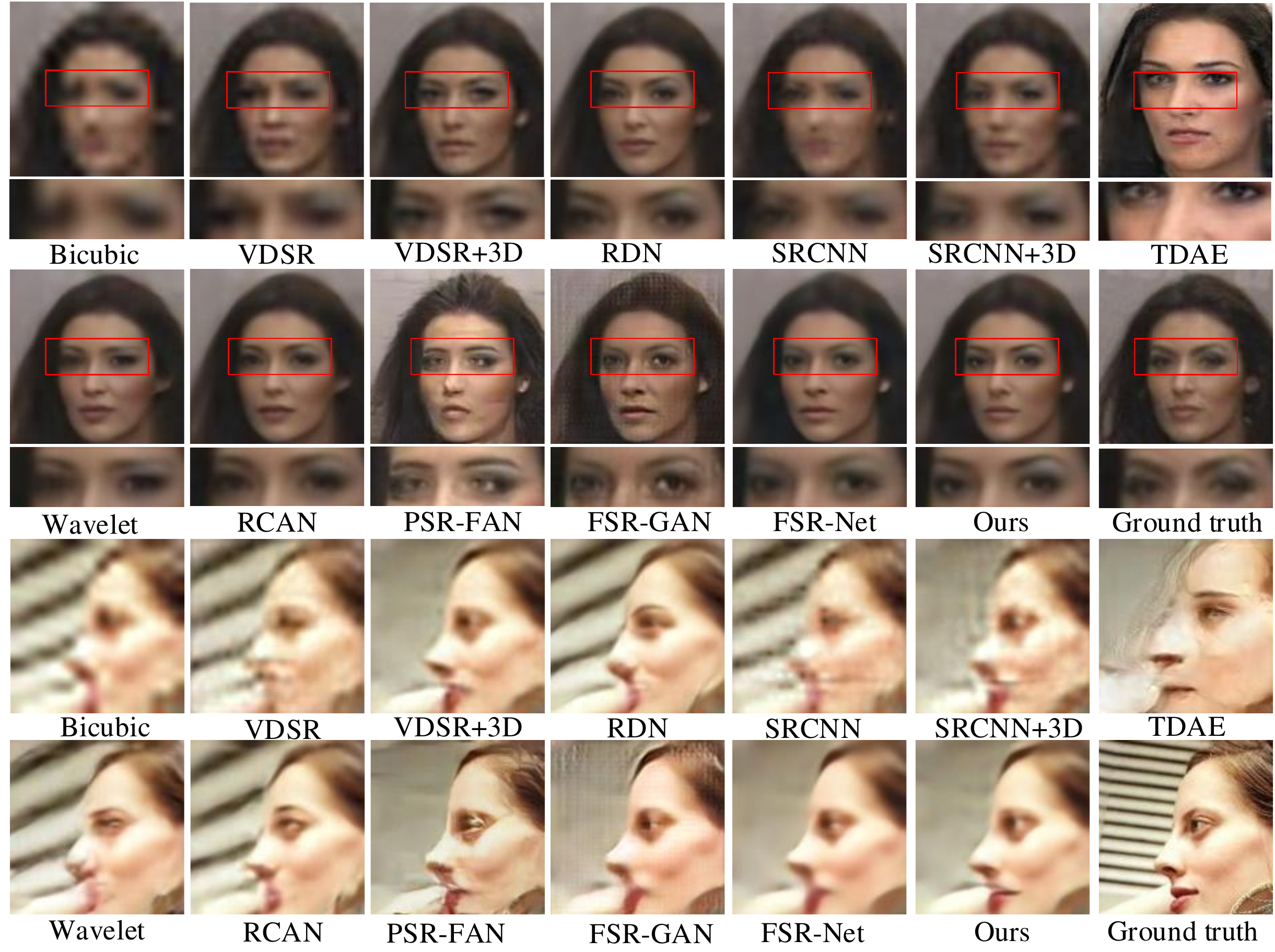}
\vspace{-5mm}
\caption{Comparison with state-of-the-art methods: magnification factors $\times$8 and the input resolution 16$\times$16. Best viewed by zooming in on the screen.}
\label{fig:fig6}
\vspace{-5mm}
\end{figure}

\begin{table}[t]
\caption{Quantitative results on the CelebA test dataset. The best results are highlighted in bold.}\label{tab:tab1}
\vspace{-2mm}
\scriptsize
%\small  \footnotesize  \normalsize
\centering
\begin{tabular}{ccccc}
\toprule[1.5pt]
%\hline
-                      & \multicolumn{4}{c}{CelebA}                                             \\ \hline
\multirow{1}{*}{Scale} & \multicolumn{2}{c}{$\times$4}            & \multicolumn{2}{c}{$\times$8}            \\ \cline{2-5} 
                       & PSNR             & SSIM            & PSNR             & SSIM            \\ \toprule[1.5pt]%\hline
Bicubic                & 27.16          & 0.8197          & 21.90          & 0.6213          \\ \hline
VDSR \cite{network15}           & 28.13          & 0.8554          & 22.76          & 0.6618          \\ \hline
RCAN \cite{network_zhang}            & 29.04          & 0.8643          & 23.26          & 0.7362          \\ \hline
RDN \cite{network_rdn}             & 29.06          & 0.8650          & 23.69          & 0.7484          \\ \hline
SRCNN \cite{network6}           & 27.57          & 0.8452          & 22.51          & 0.6659          \\ \hline
TDAE \cite{prior_yu}            & -                & -               & 20.10          & 0.5802          \\ \hline
Wavelet-SRNet \cite{network_wave}   & 28.42          & 0.8698          & 23.08          & 0.7147          \\ \hline
FSRGAN \cite{comparsion_2}   & -          & -        & 22.27        & 0.6010           \\ \hline
FSRNet \cite{comparsion_2}   & -          & -        & 22.62        & 0.6410           \\ \hline
PSR-FAN \cite{comparsion_1}   & -          & -        & 22.66        & 0.6850          \\ \hline
VDSR+3D            & 29.29          & 0.8727          & 24.66          & 0.7127          \\ \hline
Ours           & \textbf{29.69} &\textbf{0.8817} &\textbf{25.39} & \textbf{0.7551} \\ \toprule[1.5pt]
\end{tabular}
\vspace{-1mm}
\end{table}
\begin{table}[!ht]\scriptsize
\caption{Quantitative results of different large facial pose variations (e.g., left, right, and semifrontal) on the Menpo test dataset. The best results are highlighted in bold.}\label{tab:tab2}
\vspace{-2mm}
%\tiny
\scriptsize
%\small  \footnotesize  \normalsize
\centering
% Please add the following required packages to your document preamble:
% \usepackage{multirow}
% \begin{tabular}{p{1.25cm}p{0.6cm}p{0.5cm}p{0.6cm}p{0.5cm}p{0.6cm}p{0.5cm}p{0.6cm}p{0.5cm}p{0.6cm}p{0.5cm}p{0.6cm}p{0.5cm}}
\begin{tabular}{p{1.5cm}p{0.65cm}p{0.90cm}p{0.65cm}p{0.90cm}p{0.65cm}p{0.90cm}p{0.65cm}p{0.90cm}p{0.65cm}p{0.90cm}p{0.65cm}p{0.90cm}}
\toprule[1.5pt]
-                     & \multicolumn{12}{c}{Menpo}                                                                                                                                                                                                 \\ \hline
Scale                 & \multicolumn{6}{c}{$\times$4}                                                                                      & \multicolumn{6}{c}{$\times$8}                                                                                      \\ \hline
\multirow{1}{*}{Pose} & \multicolumn{2}{c}{Left}          & \multicolumn{2}{c}{Right}         & \multicolumn{2}{c}{Semi-frontal}  & \multicolumn{2}{c}{Left}          & \multicolumn{2}{c}{Right}         & \multicolumn{2}{c}{Semi-frontal}  \\ \cline{2-13} 
                      & PSNR             & SSIM            & PSNR             & SSIM            & PSNR             & SSIM            & PSNR             & SSIM            & PSNR             & SSIM            & PSNR             & SSIM            \\ \toprule[1.5pt]
Bicubic               & 26.36          & 0.7923          & 26.19          & 0.7791          & 24.92          & 0.7608          & 22.09          & 0.6423          & 21.99          & 0.6251          & 20.68          & 0.5770          \\ \hline
VDSR \cite{network15}           & 26.99          & 0.8024          & 26.85          & 0.7908          & 25.63          & 0.7794          & 22.28          & 0.6315          & 22.20          & 0.6163          & 20.98          & 0.5752          \\ \hline
RCAN \cite{network_zhang}           & 27.47          & 0.8259          & 27.27          & 0.8145          & 26.11          & 0.8080          & 21.94          & 0.6543          & 21.87          & 0.6381          & 20.60          & 0.5938       \\ \hline
RDN \cite{network_rdn}            & 27.39          & 0.8263          & 27.21          & 0.8150          & 26.06          & 0.8088          & 22.30          & 0.6706          & 22.24          & 0.6552          & 21.02          & 0.6160          \\ \hline
SRCNN \cite{network6}          & 26.92          & 0.8038          & 26.74          & 0.7913          & 25.50          & 0.7782          & 22.38          & 0.6408          & 22.32          & 0.6272          & 21.08          & 0.5857          \\ \hline
TDAE \cite{prior_yu}           & -                & -               & -                & -               & -                & -               & 21.22          & 0.5678          & 20.22         & 0.5620          & 19.88          & 0.5521          \\ \hline
Wavelet-SRNet \cite{network_wave}  & 26.97          & 0.8122          & 26.81          & 0.8001          & 25.72          & 0.7945          & 21.86          & 0.6360          & 21.72          & 0.6166          & 20.57          & 0.5779          \\ \hline
FSRGAN \cite{comparsion_2} & -& -& - &- &- &-   & 23.00& 0.6326 & 22.84  & 0.6173 & 22.00 & 0.5938  \\ \hline
FSRNet \cite{comparsion_2} & -& -& - &- &- &-   & 23.56& 0.6896 & 23.43  & 0.6712 & 22.03 & 0.6382  \\ \hline
PSR-FAN \cite{comparsion_1} & -& -& - &- &- &-   & 22.04& 0.6239 & 21.89  & 0.6114 & 20.88 & 0.5711  \\ \hline
VDSR+3D           & 28.62          & 0.8439          & 28.89          & 0.8326          & 26.99          & 0.8236          & 23.45          & 0.6845          & 23.25          & 0.6653          & 21.83          & 0.6239          \\ \hline
Ours         & \textbf{28.98} & \textbf{0.8510} & \textbf{29.29} & \textbf{0.8408} & \textbf{27.29} & \textbf{0.8332} & \textbf{23.80} & \textbf{0.7071} & \textbf{23.57} & \textbf{0.6881} & \textbf{22.15} & \textbf{0.6501} \\ \toprule[1.5pt]
\end{tabular}
\vspace{-6mm}
\end{table}

%\vspace{-5}
\subsection{Datasets and Implementation Details}
CelebA \cite{celeba} and Menpo \cite{menpo} datasets are used to verify the performance of the algorithm. The training phase uses 162,080 images from the CelebA dataset. In the testing phase, 40,519 images from the CelebA test set are used along with the large-pose-variation test set from the Menpo dataset. The every facial pose test set of Menpo (left, right and semi-frontal) contains 1000 images, respectively. We follow the protocols
of existing face SR methods (e.g., \cite{comparsion_1}, \cite{comparsion_2},\cite{prior_2},\cite{deep16}) to generate the LR input by the bicubic downsampling method. The HR ground-truth images are obtained by center-cropping the facial images and then resizing them to the 128$\times$128 pixels. The LR face images are generated by downsampling HR ground-truths to 32$\times$32 pixels ($\times$4 scale) and 16$\times$16 pixels ($\times$8 scale). In our network, the ADAM optimizer is used with a batch size of 64 for training, and input images are center-cropped as RGB channels. The initial learning rate is 0.0002 and is divided by 2 every 50 epochs. The whole training process takes 2 days with an NVIDIA Titan X GPU. 

\subsection{Quantitative Results}
Quantitative evaluation of the network using PSNR and the structural similarity (SSIM) scores for the CelebA test set is listed in Table \ref{tab:tab1}. Furthermore, to analyze the performance and stability of the proposed method with respect to large face pose variations, three cases corresponding to different face poses (left, right, and semifrontal) of the Menpo test data are listed in Table \ref{tab:tab2}.

%\vspace{-2mm}
{\flushleft \textbf{CelebA Test:}} As shown in Table \ref{tab:tab1}, VDSR+3D (the basic VDSR model \cite{network15} guided by the proposed 3D facial priors) achieves significantly better results (1 dB higher than the remaining best method and 2 dB higher than the basic VDSR method in $\times$8 SR) even for the large-scale parameter methods, such as RDN and RCAN. 
It is worth noting that VDSR+3D still performs slightly worse than the proposed algorithm of SAM3D. 
These results demonstrate that the proposed 3D priors make a significant contribution to the performance improvement (average 1.6 dB improvement) of face super-resolution. In comparison with 2D priors based methods (\textit{e.g.,} FSRNet and PSR-FAN), our algorithm performs much better (2.73 dB higher than PSR-FAN and 2.78 dB higher than FSRNet).  
%
%It should be noted that ours (VDSR+) is the same as VDSR except for the extra 3D face priors as the RGB channel inputs. 

%\vspace{-2mm}
{\flushleft \textbf{Menpo Test:}} To verify the effectiveness and stability of the proposed network towards face pose variations, the quantitative results on the dataset with large pose variations are reported in Table \ref{tab:tab2}. While ours (SAM3D) is the best method superior than the others, VDSR+3D also achieves 1.8 dB improvement compared with the basic VDSR method in the $\times$4 magnification factor. Our 3D facial priors based method is still the most effective approach to boost the SR performance compared with 2D heatmaps and parsing maps priors.

%\vspace{-10mm}
\begin{figure}[t]\scriptsize
	\begin{center}
		\tabcolsep 1pt
		\begin{tabular}{@{}ccccccc@{}}
			\includegraphics[width=0.138\textwidth]{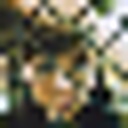} & 
			\includegraphics[width=0.138\textwidth]{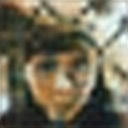} & 
			\includegraphics[width=0.138\textwidth]{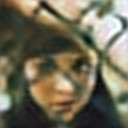} &            
			\includegraphics[width=0.138\textwidth]{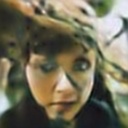} &
			\includegraphics[width=0.138\textwidth]{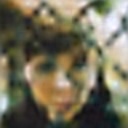} &
			\includegraphics[width=0.138\textwidth]{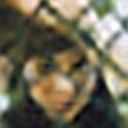} & 
			\includegraphics[width=0.138\textwidth]{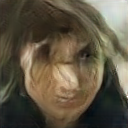}  \\
			 Bicubic &  VDSR &  VDSR+3D & RDN & SRCNN &  SRCNN+3D & TDAE  \\
			 PSNR/SSIM:& 17.02/0.4059 &19.08/0.4860  & 16.72/0.4405 &17.02/0.4158 &18.69/0.4457 &11.62/0.1666  \\
			 %\vspace{1pt}\\
			\includegraphics[width=0.138\textwidth]{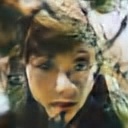}  & 
			\includegraphics[width=0.138\textwidth]{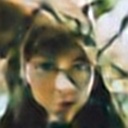} & 
			\includegraphics[width=0.138\textwidth]{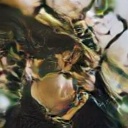} &
			\includegraphics[width=0.138\textwidth]{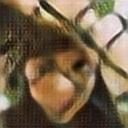}  & 
			\includegraphics[width=0.138\textwidth]{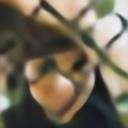} & 
			\includegraphics[width=0.138\textwidth]{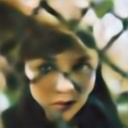} &
			\includegraphics[width=0.138\textwidth]{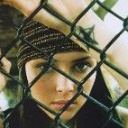} \\
			%\vspace{1pt} \\
			Wavelet &  RCAN &  PSR-FAN &   FSR-GAN &  FSR-Net &  Ours & Ground truth
			\\
			16.06/0.3732 & 16.17/0.4004  & 16.96/0.3923 & 18.39/0.4515 & 19.26/0.5043& 19.47/0.5381 & - \vspace{1pt}\\
		\end{tabular}
	\end{center}
	\vspace{-5mm}
\caption{Visual comparison with state-of-the-art methods ($\times$8). The results by the proposed method have fewer artifacts on face components (e.g., eyes, mouth, and nose).}\label{fig:fig7}
	%\vspace{-2mm}
	\label{fig:fig7}
\end{figure}
\begin{figure}[t]
\centering
\includegraphics[width=1\textwidth]{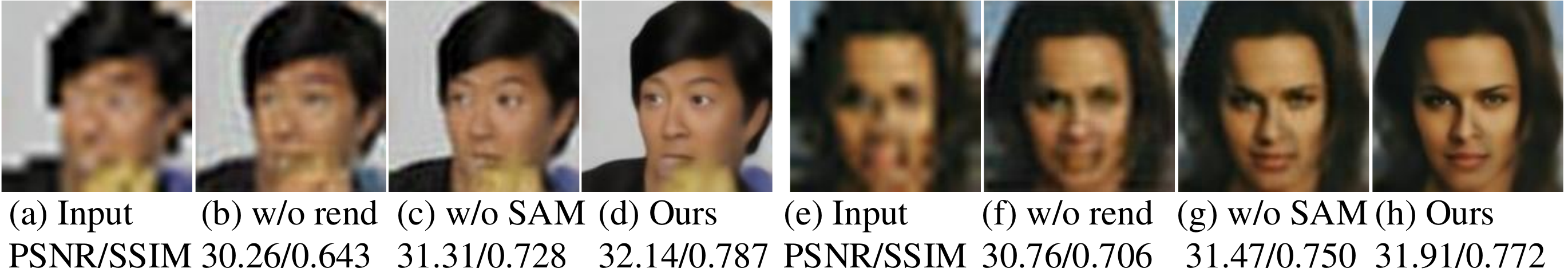}
\vspace{-6mm}
\caption{Ablation study results: Comparisons between our proposed model with different configurations, with PSNR and SSIM relative to the ground truth. (a) and (e) are the inputs. (b) and (f) are the SR results without using the rendered priors. (c) and (g) are the SR results without the Spatial Attention Module. (d) and (h) are our SR results.}\label{fig:fig8}
%\vspace{-0.4cm}
\end{figure}
%
% \begin{figure}[t]
% \centering
% \subfigure[$\times$4 scale]{
% \begin{minipage}[t]{0.48\textwidth}
% \centering
% \includegraphics[width=5.5cm]{learning_curve11.pdf}
% \end{minipage}}
% \subfigure[$\times$8 scale]{
% \begin{minipage}[t]{0.48\textwidth}
% \centering
% \includegraphics[width=5.5cm]{learning_curve22.pdf}
% \end{minipage}}
% \centering
% \vspace{-4mm}
% \caption{Test accuracy curves with different configurations along the training epochs.
% %: basic+ denotes the basic method (SRCNN and VDSR) incorporating the 3D facial priors.
% }
% \label{fig:fig9}
% \vspace{-5mm}
% \end{figure}
%
%
\begin{figure}[t]
\centering
\includegraphics[width=0.8\textwidth]{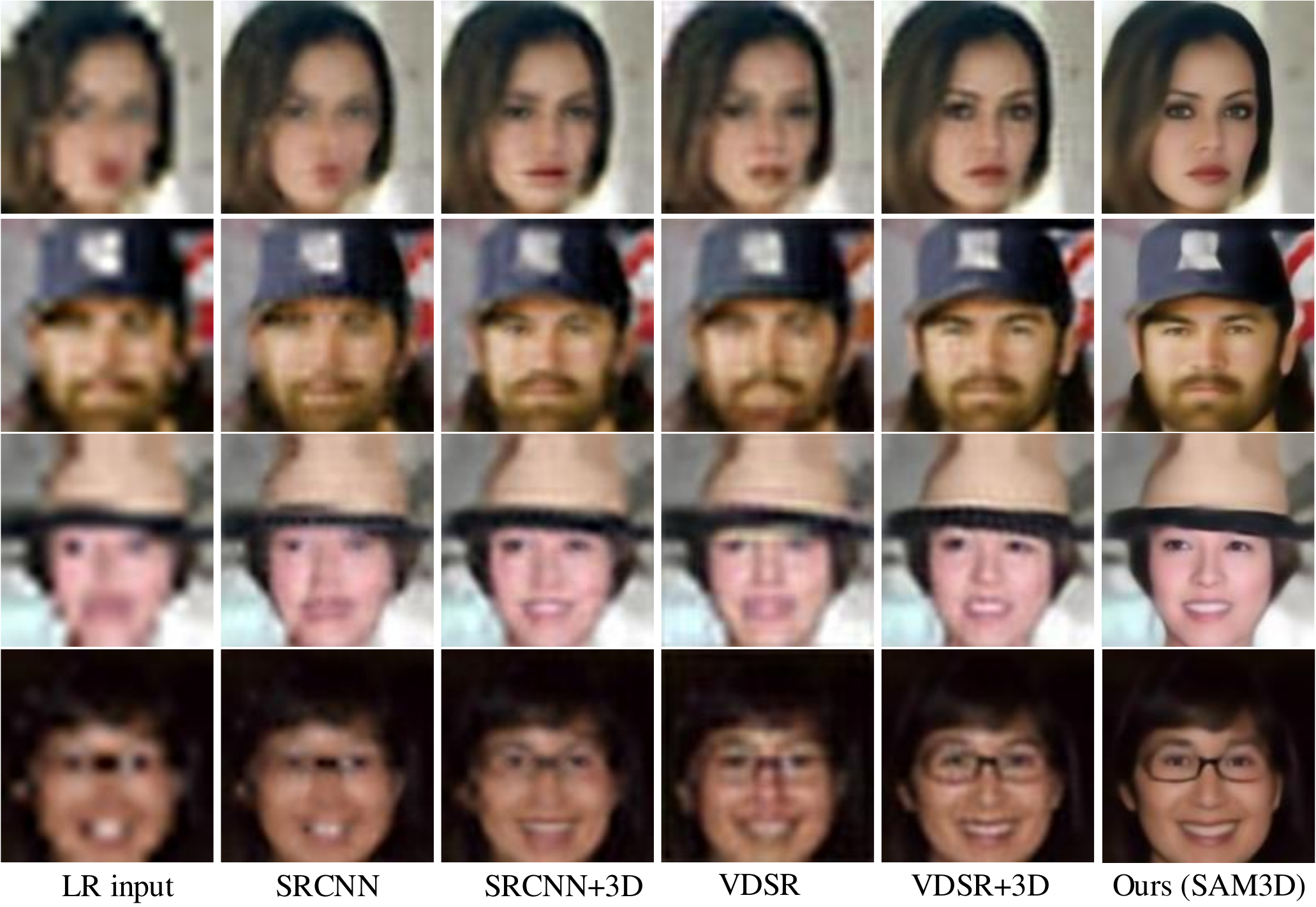}
\vspace{-4mm}
\caption{Qualitative evaluation with different ablation configurations: SRCNN+3D  and VDSR+3D denote the basic method (SRCNN and VDSR) incorporating the 3D facial priors; Ours (SAM3D) means the Spatial Attention Module incorporating the 3D facial priors. Our 3D priors enable the basic methods to avoid some artifacts around the key facial components and to generate sharper edges.}\label{fig:fig10}
%\vspace{-0.8cm}
\end{figure}
%

%\vspace{-1mm}
\subsection{Qualitative Evaluation}
%
%\textbf{Super-resolution}: 
The qualitative results of our methods at different magnifications ($\times$4 and $\times$8) are shown respectively in Figures \ref{fig:fig5} and \ref{fig:fig6}. It can be observed that our proposed method recovers clearer faces with finer component details (e.g., noses, eyes, and mouths). %\textbf{Artifacts}: 
The outputs of most methods (\textit{e.g.,} PSR-FAN, RCAN, RDN, and Wavelet-SRNet) contain some artifacts around facial components such as eyes and nose, as shown in Figures \ref{fig:fig1} and \ref{fig:fig7}, especially when facial images are partially occluded. After adding the rendered face priors, our results show clearer and sharper facial structures without any ghosting artifacts, which illustrates that the proposed 3D priors help the network understand the spatial location and the entire face structure and largely avoid the artifacts and significant distortions in facial attributes which are common in facial landmark priors, because the attention is applied merely to the distinct points of facial landmarks.
%\vspace{-2mm}
\section{Analyses and Discussions}
%\vspace{-2mm}
%
{\flushleft \textbf{Ablation Study}}: In this section, we conduct an ablation study to demonstrate the effectiveness of each module. We compare the proposed network with and without using the rendered 3D face priors and the Spatial Attention Module (SAM) in terms of PSNR and SSIM on the $\times$8 scale test data. As shown in Figure \ref{fig:fig8} (b) and (f), the baseline method without using the rendered faces and SAM tends to generate blurry faces that cannot capture sharp structures. Figure \ref{fig:fig8} (c) and (g) show clearer and sharper facial structures after adding the 3D rendered priors. By using both SAM and 3D priors, the visual quality is further improved in Figure \ref{fig:fig8} (d) and (h). The quantitative comparisons between (VDSR, our VDSR+3D, and our SAM3D) in Tables \ref{tab:tab1} and \ref{tab:tab2} also illustrate the effectiveness of the proposed rendered priors and the spatial attention module.

To verify the advantage of 3D facial structure priors in terms of the convergence and accuracy, three different configurations are designed: basic methods (\textit{i.e.,} SRCNN \cite{network6} and VDSR \cite{network15}); basic methods incorporating 3D facial priors (\textit{i.e.,} SRCNN+3D and VDSR+3D); the proposed method using the Spatial Attention Module and 3D priors (SAM3D). The validation accuracy curve of each configuration along the epochs is plotted to show the effectiveness of each block. The priors are easy to insert into any network. They only marginally increase the number of parameters, but significantly improve the accuracy and convergence of the algorithms as shown in Supplementary Fig.3. The basic methods of SRCNN and VDSR incorporating the facial rendered priors tend to avoid some artifacts around key facial components and generate sharper edges compared to the baseline methods without the facial priors. By adding the Spatial Attention Module, it helps the network better exploit the priors and easily enables to generate sharper facial structures as shown in Figure \ref{fig:fig10}.

{\flushleft \textbf{Results on Real-World Images}}: For real-world LR images, we provide the quantitative and qualitative analysis on 500 LR faces from the WiderFace (x4) dataset in Supplementary Tab.1 and Fig.1.

{\flushleft \textbf{Model Size and Running Time}}: We evaluate the proposed method and STOA SR methods on the same server with an Intel Xeon W-2123 CPU and an NVIDIA TITAN X GPU. Our proposed SAM3D, embedded with 3D priors, are more lightweight and less time-consuming, shown in Supplementary Fig.2.  

%as well as the VDSR$+$3D,
% while still achieving the best performance even than recent SOTA SR methods (\textit{e.g.,} RCAN and RDN) and face priors based SR methods (\textit{e.g.,} FSRNet and PSRFAN), shown in Supplementary Fig.2.

% %with a larger scale of parameters.
% %\subsection{Test Running Time}

% In addition, as shown in Figure \ref{fig:fig11} (b), our proposed method Spatial Attention Module incorporating 3D priors (SAM3D) and VDSR+3D improve PSNR for scale factor $\times$8 on the dataset CelebA in comparison to the state-of-the-art methods. Our methods outperform the other approaches by a large margin while maintaining comparable run times with face SR methods with 2D priors. Our test running time includes the time required for the ResNet-50. 

%\vspace{-2mm}
\section{Conclusions}
%\vspace{-2mm}
In this paper, we proposed a face super-resolution network that incorporates the novel 3D facial priors of rendered faces and multi-dimensional knowledge. In the 3D rendered branch, we presented a face rendering loss to encourage a high-quality guided image providing clear spatial locations of facial components and other hierarchical information (\textit{i.e.,} expression, illumination, and face pose). Compared with the existing 2D facial priors whose attentions are focused on the distinct points of landmarks which may result in face distortions, our 3D priors are explicit, visible and highly realistic, and can largely decrease the occurrence of face artifacts. To well exploit 3D priors and consider the channel correlation between priors and inputs, we employed the Spatial Feature Transform and Attention Block. The comprehensive experimental results have demonstrated that the proposed method achieves superior performance and largely decreases artifacts in contrast with the SOTA methods.

\subsection*{Acknowledgement}
This work is supported by the National Key R$\&$D Program of China under Grant 2018AAA0102503, Zhejiang Lab (NO.2019NB0AB01), Beijing Education Committee Cooperation Beijing Natural Science Foundation (No.KZ201910005007), National Natural Science Foundation of China (No.U1736219) and Peng Cheng Laboratory Project of Guangdong Province PCL2018KP004.

\bibliographystyle{splncs04}
\bibliography{eccv2020submission}
\end{document}